\documentclass[10pt,twocolumn,letterpaper]{article}

\usepackage[pagenumbers]{cvpr} 

\usepackage{algorithm, algpseudocode}
\usepackage{xscommand}
\usepackage{enumitem}
\usepackage{pifont}
\usepackage{wrapfig}
\usepackage{stfloats}
\usepackage{graphicx}

\usepackage{subcaption}
\definecolor{cvprblue}{rgb}{0.21,0.49,0.74}
\usepackage[pagebackref,breaklinks,colorlinks,citecolor=cvprblue]{hyperref}

\def\paperID{5882} 
\def\confName{CVPR}
\def\confYear{2024}

\title{Rethinking Mixup for Improving the Adversarial Transferability}

\author{Xiaosen Wang\\
Huawei Singular Security Lab\\
{\tt\small xiaosen@hust.edu.cn}
\and
Zeyuan Yin\\
Mohamed bin Zayed University of AI\\
{\tt\small zeyuan.yin@mbzuai.ac.ae}
}

\begin{document}
\maketitle

\begin{abstract}

Mixup augmentation has been widely integrated to generate adversarial examples with superior adversarial transferability when immigrating from a surrogate model to other models. However, the underlying mechanism influencing the mixup's effect on transferability remains unexplored. In this work, we posit that the adversarial examples located at the convergence of decision boundaries across various categories exhibit better transferability and identify that \textit{Admix} tends to steer the adversarial examples towards such regions. However, we find the constraint on the added image in \textit{Admix} decays its capability, resulting in limited transferability. To address such an issue, we propose a new input transformation-based attack, called \textbf{M}ixing the \textbf{I}mage but \textbf{S}eparating the gradien\textbf{T} (\name). Specifically, \name randomly mixes the input image with a randomly shifted image and separates the gradient of each loss item for each mixed image. To counteract the imprecise gradient, \name calculates the gradient on several mixed images for each input sample. Extensive experimental results on the ImageNet dataset demonstrate that \name outperforms existing SOTA input transformation-based attacks with a clear margin on both Convolutional Neural Networks (CNNs) and Vision Transformers (ViTs) w/wo defense mechanisms, supporting \name's high effectiveness and generality.

\end{abstract}

\section{Introduction}


Adversarial examples, in which the imperceptible perturbations can mislead the model predictions~\cite{szegedy2014intriguing,goodfellow2015explaining}, reveal the extreme vulnerability of deep neural networks (DNNs)~\cite{simonyan2015very,he2016deep}. It has also brought an immense threat to real-world DNNs-based applications, such as face recognition~\cite{wen2016discriminative,wang2018cosface}, autonomous driving~\cite{chen2015deepdriving,eykholt2018robust}, \etc. Recently, with the increasing interest in adversarial examples, researchers have proposed tremendous adversarial attacks, such as white-box attacks~\cite{kurakin2017adversarial,madry2018towards}, score-based attacks~\cite{ilyas2018black,guo2019simple}, decision-based attacks~\cite{brendel2018decision,wang2022triangle} and transfer-based attacks~\cite{dong2018boosting,wei2019transferable,wang2023rethinking}. 

Among various adversarial attacks, transfer-based attacks utilize the transferability of adversarial examples generated on the surrogate model to attack other unknown models~\cite{liu2017delving}. Since transfer-based attacks do not access the target model, they are applicable to attack real-world models. However, existing adversarial attacks often exhibit superior white-box attack performance but poor transferability. To this end, numerous attacks have been proposed to boost the adversarial transferability~\cite{xie2019improving,dong2019evading,lin2020nesterov,gao2020patch,wang2023structure,wu2021improving,wang2021boosting,zhang2022improving,zhang2022enhancing}. 

\begin{figure}
    \centering
    \includegraphics[width=\linewidth]{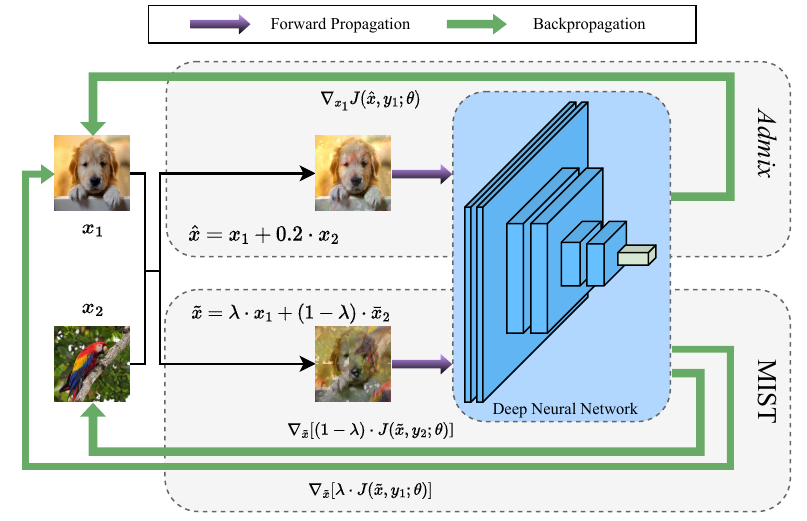}
    \caption{Illustration of the gradient calculation in \textit{Admix} and \name. The transformed image of \textit{Admix} is visually similar to the original input image $x_1$, while the one of \name is a linear combination of two input images $x_1$ and $\bar{x}_2$. in which $\bar{x}_2$ indicates the randomly shifted image of $x_2$.}
    \label{fig:illustration}
    \vspace{-1em}
\end{figure}

Input transformation-based attacks, which transform the input image for gradient calculation, exhibit outstanding effectiveness in boosting the transferability and have attracted increasing attention~\cite{xie2019improving,ge2023boosting,lin2020nesterov,wang2021admix,wang2023boosting}. As shown in Fig.~\ref{fig:illustration}, \textit{Admix}~\cite{wang2021admix} achieves outstanding performance among existing input transformation-based attacks by adding a small portion of the image from other categories to the input image for gradient calculation. In this work, we postulate and empirically validate that the adversarial examples located at the intersection of decision boundaries across various categories are more transferable using a toy two-dimensional classification task. By leveraging information from other categories, \textit{Admix} effectively guides adversarial examples toward these convergent points of decision boundaries, which is attributed to its excellent transferability.




Nevertheless, \textit{Admix} adopts a small portion of the image to prevent an imprecise gradient introduced by the overload of information from different categories. Such a constraint might limit the capability to steer adversarial examples towards the intersection of decision boundaries, which might limit the transferability. In this work, we identify that separating the gradient of each loss item can overcome this limitation. Based on this finding, we propose a novel input transformation-based attack, called \textbf{M}ixing the \textbf{I}mage but \textbf{S}eparating the gradien\textbf{T} (\name), to further boost the adversarial transferability. As depicted in Fig.~\ref{fig:illustration}, \name distinctively mixes two input images without prioritizing one over the other.
Besides, \name employs a random shift in one image to augment the diversity of the transformed images. To prevent the imprecise information in the gradient from decaying the attack performance, \name generates $N$ mixed images for each input sample and calculates the average gradient across these images.


We summarize our contributions as follows:
\begin{itemize}[leftmargin=*,noitemsep,topsep=2pt]
\item We hypothesize that the adversarial examples located at the intersection of decision boundaries across various categories exhibit enhanced transferability. We provide a toy example to validate this assumption and identify that \textit{Admix} guides the adversarial examples towards such region, leading to better transferability.
\item We find that the constraint on the added image in \textit{Admix} limits the transferability and separating the gradient of each loss item can address such limitation. Based on this finding, we propose a new input transformation-based attack called \name, which randomly mixes the input image with a randomly shifted image. To counteract the imprecise gradient, \name calculates the gradient on $N$ mixed images for each input sample. With a single forward propagation, \name can calculate the gradient \wrt two mixed images, making it more efficient than \textit{Admix}.
\item Extensive experiments on ImageNet dataset demonstrate that \name outperforms the winner-up approach among existing input transformation-based attacks with an average margin of $10.32\%$ on either CNN-based or transformer-based models under various settings, showing its superiority and generality.
\end{itemize}

\section{Related Work}
In this section, we provide a brief overview of existing adversarial attacks and defenses.

\subsection{Adversarial Attacks}
Since Szegedy~\etal~\cite{szegedy2014intriguing} identified the vulnerability of DNNs against adversarial examples, numerous adversarial attacks have been proposed, such as white-box attacks~\cite{goodfellow2015explaining,kurakin2017adversarial,madry2018towards}, score-based attacks~\cite{ilyas2018black,guo2019simple}, decision-based attacks~\cite{brendel2018decision,wang2022triangle} and transfer-based attacks~\cite{liu2017delving,dong2018boosting,xie2019improving,wang2021enhancing,ge2023boosting}. Transfer-based attacks have been widely investigated recently since they do not access the target model, making them applicable in the physical world. 

MI-FGSM~\cite{dong2018boosting} introduces momentum into I-FGSM~\cite{kurakin2017adversarial} to stabilize the update direction and boost transferability, which has motivated several momentum-based approaches, \eg, NI-FGSM~\cite{lin2020nesterov}, VMI-FGSM~\cite{wang2021enhancing}, EMI-FGSM~\cite{wang2021boosting}, \etc. Ensemble attack~\cite{liu2017delving,xiong2022stochastic}, which simultaneously attacks multiple deep models with different architectures, is also effective in crafting transferable adversarial examples. Besides, some regularizers~\cite{zhou2018transferable,wu2021improving} on the intermediate features or attention heatmaps can improve the transferability. Researchers also investigate the architecture of the victim model~\cite{li2020learning,wu2020skip} or utilize the generative models to craft more transferable adversarial examples~\cite{poursaeed2018generative,naseer2021generating}.

Aside from the above approaches, input transformation-based attacks, which randomly transform the input image before gradient calculation, have shown remarkable effectiveness in enhancing the transferability and excellent compatibility with any existing transfer-based adversarial attacks. For instance, DIM~\cite{xie2019improving} randomly resizes the input image and adds padding to obtain a transformed image with a fixed size before calculating the gradient. TIM~\cite{dong2019evading} convolves the gradient of the input image with a Gaussian kernel to approximate the average gradient on several translated images. SIM~\cite{lin2020nesterov} calculates the average gradient on ensemble scaled images using different scale factors. \textit{Admix}~\cite{wang2021admix} adds a small portion of the image from other categories to the input image for gradient calculation, in which SIM is a particular case of \textit{Admix} without the image from other categories. AITL~\cite{yuan2022adaptive} trains a deep neural network to adaptively predict the transformations for the given image among several image transformations. SSA~\cite{long2022frequency} adds Gaussian noise and randomly scales the input image in the frequency domain.

In this work, we validate that \textit{Admix} steers the adversarial examples towards the intersection of decision boundaries across various categories, resulting in more transferable adversarial examples. However, the constraint on the added image decays its capability, which limits the transferability. Based on this finding, we propose a new input transformation-based attack to effectively utilize the information from other categories for better transferability.

\subsection{Adversarial Defenses}
Numerous adversarial defenses have been proposed recently to mitigate the threat of adversarial examples. For example, adversarial training~\cite{goodfellow2015explaining,madry2018towards,tramer2018ensemble,wong2020fast} injects the adversarial examples into the training set to effectively enhance the model robustness but also introduces huge computational cost and degrades the performance on benign data. On the other hand, several approaches struggle to eliminate the adversarial perturbation through various input pre-processing strategies, such as JPEG compression~\cite{guo2018countering}, random Resizing and Padding (R\&P)~\cite{xie2018mitigating}, or training a deep denoiser to purify the input image, namely High-level representation Guided Denoiser (HGD)~\cite{liao2018defense}, JPEG-based Feature Distillation (FD)~\cite{liu2019feature}, Neural Representation Purification (NRP)~\cite{naseer2020a}, \etc. In contrast to the above empirical defenses, certified defenses offer provable defense in a given radius, such as Interval Bound Propagation (IBP)~\cite{gowal2019scalable}, CROWN~\cite{zhang2018efficient} and Randomized Smoothing (RS)~\cite{cohen2019certified}.

\section{Approach}
In this section, we rethink how to boost adversarial transferability using a toy example and detail our \name.
\subsection{Rethinking Adversarial Transferability}
\label{sec:approach:rethinking}
It is widely recognized that the objective of model training is to align with or approximate the inherent distribution of a given dataset. Considering the static nature of the data distribution, we can posit the following hypothesis regarding the decision boundary of various models with superior performance:
\begin{assumption}[Similarity of decision boundary]
    Given two distinct models $f_1$ and $f_2$, which are trained on an identical dataset and yield comparable results, their respective decision boundaries exhibit a degree of similarity.
\end{assumption}
Intuitively, the better performance achieved by the models, the higher degree of similarity in approximating the data distribution. In other words, their decision boundaries tend to be more aligned. This phenomenon underpins the rationale behind the transferability of adversarial examples across different models. Thus, the adversarial examples far away from the decision boundary around the ground-truth category tend to be more transferable.

\begin{figure}
    \centering
    \includegraphics[width=0.85\linewidth]{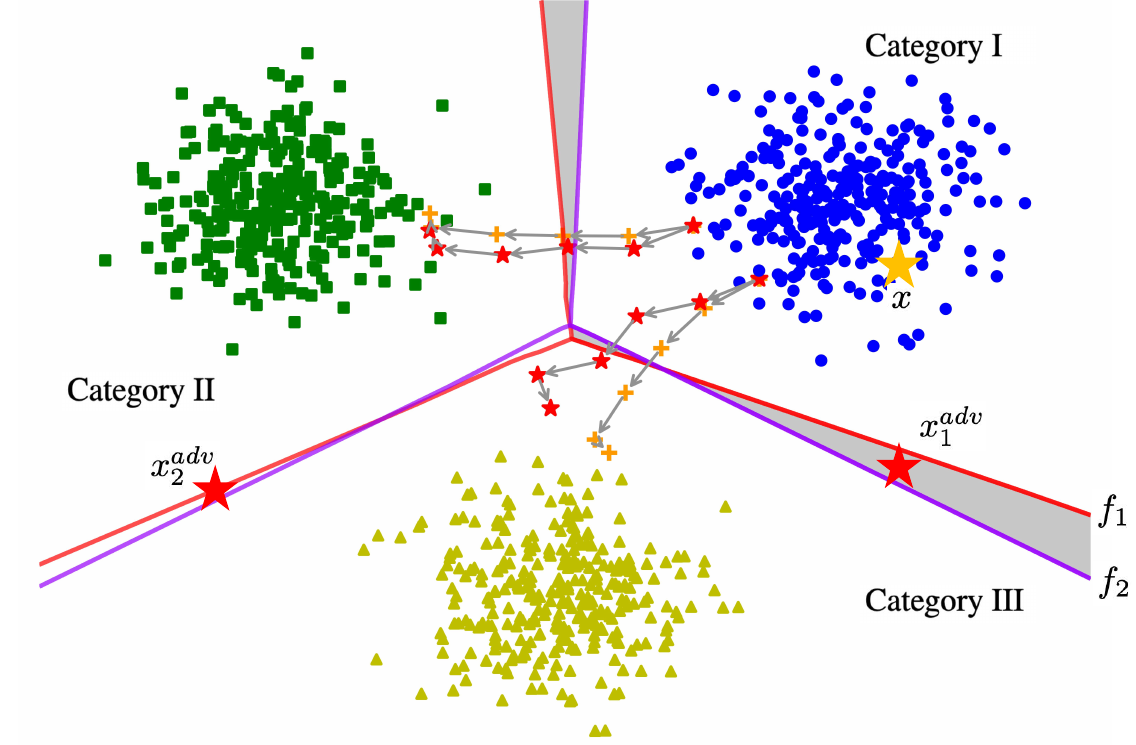}
    \caption{Visualization of decision boundaries for distinct models $f_1$ and $f_2$ trained on the two-dimensional dataset, data point $x$ and adversarial examples $x_1^{adv}$ and $x_2^{adv}$. The orange and red trajectories are the optimization paths of MI-FGSM and \textit{Admix}.}
    \label{fig:vis_motivation}
    \vspace{-0.5em}
\end{figure}

To elucidate this concept more clearly, we devise a two-dimensional dataset encompassing three distinct categories. Then we train two linear models, denoted as $f_1$ and $f_2$, both of which demonstrate exceptional performance. The decision boundaries of these models are graphically visualized in Fig.~\ref{fig:vis_motivation}. For the data point $x$ in the Category \RomanNum{1}, we observe that when adversarial examples are generated on model $f_1$, both $x_1^{adv}$ and $x_2^{adv}$ emerge as viable solutions. Though they both cross over the decision boundaries of model $f_1$, $x_2^{adv}$ possesses the capability to deceive model $f_2$ while $x_1^{adv}$ lacks this attribute. In other words, $x_2^{adv}$ is more transferable than $x_1^{adv}$. From this sample, we can observe that regardless of variations in decision boundaries across different models, a well-trained linear model is inherently incapable of classifying data points in the intersecting regions between Categories \RomanNum{2} and \RomanNum{3} as Category \RomanNum{1}. Thus,  the adversarial examples for Category \RomanNum{1} positioned at the intersection of the decision boundaries for Categories \RomanNum{2} and \RomanNum{3} exhibit the highest degree of adversarial transferability.

Inspired by mixup~\cite{zhang2018mixup}, \textit{Admix}~\cite{wang2021admix} has achieved superior adversarial transferability by adopting the information of images from diverse categories. Here we delve into the underlying factors contributing to its remarkable improvement in adversarial transferability using the above toy example. Specifically, we adopt MI-FGSM~\cite{dong2018boosting} and \textit{Admix}~\cite{wang2021admix} to conduct untargeted attacks for a randomly sampled data point in Category \RomanNum{1}. In this process, we disregard the scale factor to solely focus on the impact of the mixup operation, which calculates the gradient as follows:
\begin{equation}
    \bar{g}_{t+1} = \frac{1}{N}\sum_{i=1}^N \nabla_{x_{t}^{adv}}J( x_{t}^{adv} + \eta \cdot x_i,y;\theta),
    \label{eq:admix}
\end{equation}
where $N=5$ is the number of sampled data points and $\eta=0.2$ is the mix strength.
In contrast to MI-FGSM, as depicted in Fig~\ref{fig:vis_motivation}, the introduction of information from other categories via \textit{Admix} notably steers the adversarial examples towards the convergence point of decision boundaries for Categories \RomanNum{2} and \RomanNum{3}. This strategic directionality of \textit{Admix} culminates in its superior effectiveness in facilitating adversarial transferability. However, it's important to note that incorporating an excess of information from other categories can lead to less accurate gradients. This reduction in gradient precision negatively impacts the efficacy of white-box attacks, as evidenced in Tab. 1 of~\cite{wang2021admix}. To mitigate this issue, \textit{Admix} employs a mix strength $\eta=0.2$, ensuring that the proportion of the added image remains relatively minor. This restriction, however, sometimes hinders \textit{Admix}'s ability to consistently guide adversarial examples to the critical intersection of decision boundaries for Categories \RomanNum{2} and \RomanNum{3} (as shown in the above trajectories in Fig.~\ref{fig:vis_motivation}), which limits the adversarial transferability.

It is noted that for the multi-class task and non-linear models, we can focus on the decision boundary in the vicinity of the input sample, which retains the same properties as the above example.

\subsection{MIST}

As discussed in Sec.~\ref{sec:approach:rethinking}, \textit{Admix} imposes a constraint on the mix strength to address the problem where an overload of information from different categories leads to imprecise gradients. This limitation, however, inadvertently restricts the potential for adversarial transferability. This is because, in certain scenarios, a greater infusion of information from other categories is necessary to effectively direct adversarial examples towards the intersection point of decision boundaries for other categories. In this paper, we aim to explore how to lift this restriction on mix strength in \textit{Admix} without the side effect of inducing gradient inaccuracies.

Let us go back to mixup~\cite{zhang2018mixup}, which combines two input samples $(x_1, y_1)$ and $(x_2, y_2)$ in the following manner:
\begin{equation}
    \tilde{x} = \lambda \cdot x_1 + (1-\lambda) \cdot x_2, \quad \tilde{y} = \lambda \cdot y_1 + (1-\lambda) \cdot y_2,
    \label{eq:mixup}
\end{equation}
where $\lambda$ is a factor within the range of $[0,1]$. Taking the cross-entropy loss function as an example, the loss can be computed as:
\begin{equation}
    J(\tilde{x}, \tilde{y};\theta) = \underbrace{\lambda \cdot J(\tilde{x}, y_1;\theta)}_{\mathrm{Item \ \RomanNum{1}}} + \underbrace{(1 - \lambda) \cdot J(\tilde{x},y_2;\theta)}_{\mathrm{Item \ \RomanNum{2}}}.
    \label{eq:mixup_loss}
\end{equation}
In Eq.~\eqref{eq:mixup_loss}, Item \RomanNum{2} introduces much imprecise gradient \wrt category $y_2$ for the input sample $x_1$. \textit{Admix} simply eliminates Item \RomanNum{2} and sets  $\lambda=0.2$ to ensure that $x_2$ does not excessively impact the gradient calculation in Item \RomanNum{1}. 

\begin{figure}[tb]
    \centering
    \includegraphics[width=0.9\linewidth]{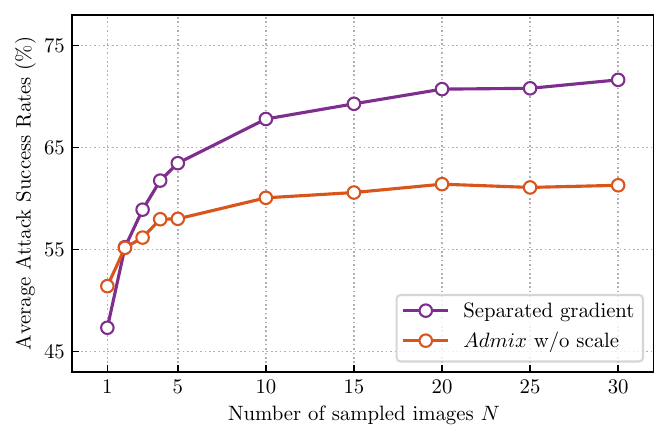}
    \vspace{-1em}
    \caption{Average attack success rate ($\%$) of separated gradient using Eq.~\eqref{eq:mist} and \textit{Admix} without scale, where the adversarial examples are generated on ResNet-18.}
    \label{fig:mix_value}
    \vspace{-1em}
\end{figure}

In this work, we argue that \textit{mixing two images should not prioritize one image over the other to achieve better transferability}. By emphasizing the equal representation of both categories in the mixed image, we can steer the adversarial examples towards the intersection of decision boundaries for different categories. To address the issue of imprecise gradients, we propose to separate the two items in Eq.~\eqref{eq:mixup_loss} when calculating the gradient \wrt each input sample as:
\begin{equation}
    \begin{aligned}
    \nabla_{x_1} J(\tilde{x}, \tilde{y};\theta) &= \nabla_{\tilde{x}}[\lambda \cdot J(\tilde{x}, y_1;\theta)],\\
    \nabla_{x_2} J(\tilde{x}, \tilde{y};\theta) &= \nabla_{\tilde{x}}[(1-\lambda) \cdot J(\tilde{x}, y_2;\theta)].
    \end{aligned}
    \label{eq:mist}
\end{equation}
To validate whether the separated gradient calculation can indeed resolve the issue of gradient imprecision and enhance adversarial transferability, we conduct an empirical experiment using Eq.~\eqref{eq:mist} and \textit{Admix} without scale. Specifically, we generate adversarial examples on ResNet-18~\cite{he2016deep} and evaluate them on seven unknown target models,
ResNet-101~\cite{he2016deep}, 
ResNeXt-50~\cite{xie2017aggregated}, 
DenseNet-121~\cite{huang2017densely}, 
ViT~\cite{dosovitskiy2020vit},
PiT~\cite{heo2021rethinking},
Visformer~\cite{chen2021visformer}, and
Swin~\cite{liu2021swin}, as detailed in Sec.~\ref{sec:exp:setting}. As shown in Fig.~\ref{fig:mix_value}, employing Eq.\eqref{eq:mist} leads to significantly improved transferable adversarial performance compared to the traditional \textit{Admix} approach, lending credence to our hypothesis. 

Based on the above analysis, we propose a new input transformation-based attack, called \textbf{M}ixing the \textbf{I}mage but \textbf{S}eparating the gradien\textbf{T} (\name). Specifically, \name adopts Eq.~\eqref{eq:mixup} to generate the mixed image while simultaneously utilizing Eq.~\eqref{eq:mist} to compute the gradient pertinent for each sample. Note that \name conducts one forward propagation when calculating the gradient of two input images, making it more efficient than \textit{Admix}.
Moreover, \name incorporates a random shift of the image $x_2$ to augment the diversity of mixed images. To mitigate the variability introduced by the stochastic mixup process, \name generates $N$ mixed images for each input sample followed by the computation of the average gradient across these images. 

\section{Experiments}
In this section, we conduct extensive experiments on ImageNet dataset to validate the effectiveness of \name.
\subsection{Experimental Settings}
\label{sec:exp:setting}
\textbf{Dataset.}
We conduct extensive experiments on 1,000 randomly sampled images belonging to 1,000 categories from the ILSVRC 2012 validation set~\cite{Russa2015imagenet}, which all the corresponding models can correctly classify.

\textbf{Models.}
To thoroughly evaluate \name, we adopt four CNN-based networks, \ie, ResNet-18, ResNet-101~\cite{he2016deep}, 
ResNeXt-50~\cite{xie2017aggregated}, 
DenseNet-121~\cite{huang2017densely}, as well as four transformer-based networks, namely ViT~\cite{dosovitskiy2020vit},
PiT~\cite{heo2021rethinking},
Visformer~\cite{chen2021visformer},
Swin~\cite{liu2021swin}. We also study several SOTA defense approaches, namely the top-3 submissions in NIPS 2017 defense competition, \ie, HGD~\cite{liao2018defense}, R\&P~\cite{xie2018mitigating} and NIPS-r3~\footnote{\url{https://github.com/anlthms/nips-2017/tree/master/mmd}}, three extra input pre-processing strategies, \ie, JPEG~\cite{guo2018countering}, FD~\cite{liu2019feature}, NRP~\cite{naseer2020a} and one certified defense RS~\cite{cohen2019certified}.

\textbf{Baseline.}
We compare \name with five widely-adopted input transformation-based attacks, namely TIM~\cite{dong2019evading}, DIM~\cite{xie2019improving}, SIM~\cite{lin2020nesterov}, \textit{Admix}~\cite{wang2021admix}, and SSA~\cite{long2022frequency}. For a fair comparison, all the input transformations are integrated into MI-FGSM~\cite{dong2018boosting}, which is aligned with previous works.

\textbf{Parameter Settings.}
We follow the parameter settings in \textit{Admix}~\cite{wang2021admix} with the maximum perturbation $\epsilon = 16$, the number of iteration $T = 10$, the step size $\alpha=\epsilon/T=1.6$, and the decay factor $\mu = 1.0$ for MI-FGSM.
For TIM, we utilize the Gaussian kernel with the size of $7\times7$.
For DIM, we set the transformation probability $p = 0.5$, and the resize rate $\rho = 1.1$. For SIM, we adopt the number of scaled copies $m = 5$ with the scaled factor $\gamma_i = 1/2^i$.
For \textit{Admix}, we follow the same scale setting in SIM and choose the number of sampled images $m_2 = 3$ with the mix strength $\eta = 0.2$. For SSA, we adopt the tuning factor $\rho=0.5$ and the standard deviation as the perturbation budget $\epsilon$.
For \name, we set the number of sampled images $N = 30$ with the dynamic mix strength $\lambda \in \left[0.2, 0.8\right]$.

\begin{figure*}[t]
    \centering
    \includegraphics[width=\linewidth]{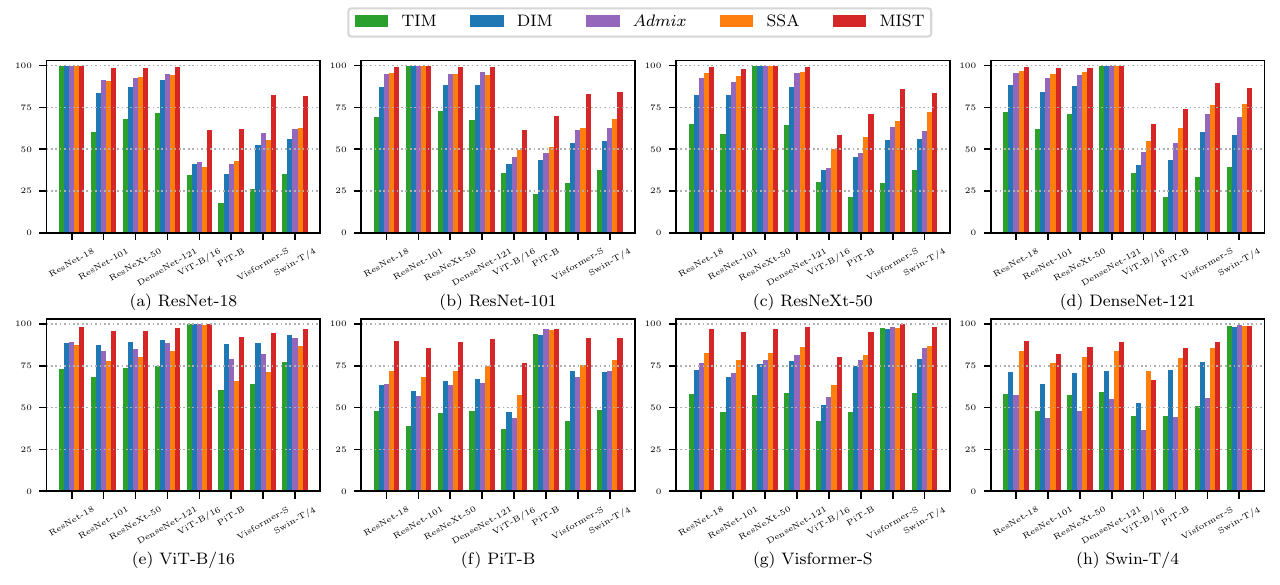}
    \vspace{-2.em}
    \caption{Attack success rates (\%) on eight standardly trained models with various single input transformations, where the adversarial examples are generated on each model using the corresponding attacks.}
    \label{fig:single_input}
    \vspace{-1em}
\end{figure*}

\subsection{Evaluation on Single Model}
\label{sec:single_model}
To verify the effectiveness of \name, we adopt various single input transformation-based attacks and further integrate \name into them to craft adversarial examples on a single model. The adversaries are crafted on the eight standardly trained models respectively, and tested on the remaining seven models to evaluate the black-box transferability.

\subsubsection{Evaluation on Single Input Transformation.}
We first adopt single input transformation to attack the deep models and report the attack success rates in Fig.~\ref{fig:single_input}, \ie, the misclassification rates of the victim model on the generated adversarial examples. 

Under the white-box setting, the baselines achieve near $100\%$ attack success rate on most models except for PiT-B. On this model, \textit{Admix} achieves the best white-box attack performance ($\sim 95\%$) among the four baselines, indicating that mixing the information from other categories is of great benefit to boost the adversarial attack. By contrast, the attack success rates of \name consistently reach almost $100\%$ on all eight models, supporting its reasonable design and showing its excellent generality for various architectures.

Under the black-box setting, among four baselines, \textit{Admix} or SSA exhibits the best attack performance when generating adversarial examples on CNN-based models. However, DIM surprisingly exhibits better attack performance when the adversarial examples are crafted on transformer-based models. This indicates the significance of thoroughly validating the attack performance on different architectures. When generating the adversarial examples on eight different models, \name exhibits outstanding transferability compared with the baselines. In general, \name achieves an attack success rate of $87.90\%$ on average, which is much higher than the best baseline \textit{Admix} ($69.70\%$) and SSA ($76.11\%$). On all models, \name outperforms the winner-up approach with an average margin of $10.32\%$. 


In summary, \name performs much better than all the baselines under white-box and black-box settings on either CNN-based or transformer-based models, demonstrating its high effectiveness in boosting adversarial attacks and excellent generality for various architectures.

\begin{figure*}[tb ]
    \centering
    \includegraphics[width=\linewidth]{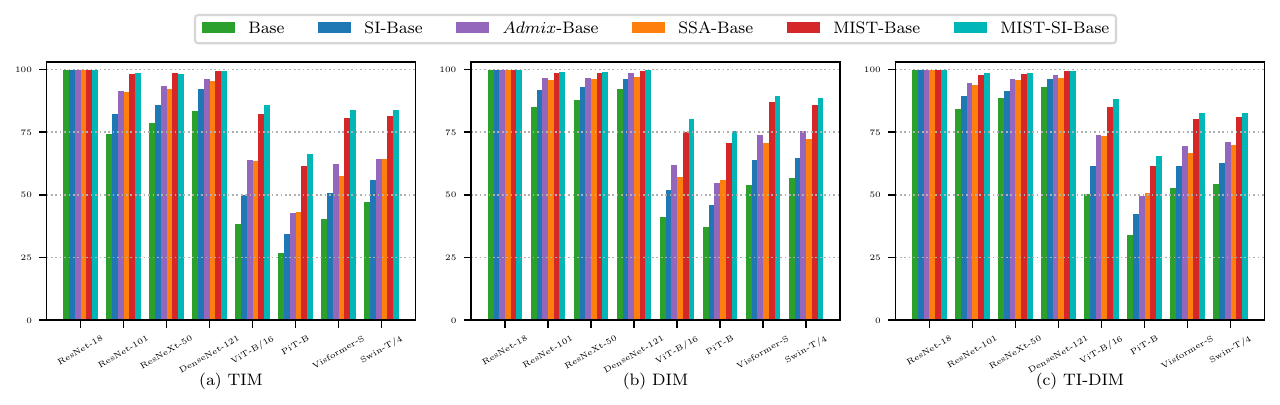}
    \vspace{-2.5em}
    \caption{Attack success rates (\%) on eight standardly trained models using combined input transformations. The adversarial examples are crafted on ResNet-18 using the corresponding attacks. Here \textbf{Base} indicates the basic input transformations, \ie, TIM, DIM, and TI-DIM.}
    \label{fig:combined_input}
\end{figure*}

\begin{table*}[t]
\centering
\resizebox{0.9\linewidth}{!}{
\begin{tabular}{@{}lccccccccc@{}}
\toprule
Ensemble                     & Attack                & ResNet-18     & ResNet-101    & ResNeXt-50    & DenseNet-121  & ViT           & PiT           & Visformer     & Swin          \\ \midrule
\multirow{4}{*}{CNNs}         & \textit{Admix}        & ~~99.8*       & ~~99.9*       & ~~99.9*       & 100.0*        & 77.2          & 81.3          & 92.3          & 91.4          \\
                             & MIST                  & ~~99.8*       & ~~99.7*       & ~~99.9*       & ~~99.9*       & \textbf{87.1} & \textbf{90.4} & \textbf{97.4} & \textbf{95.9} \\ \cdashline{2-10}\noalign{\vskip 0.5ex}
                             & \textit{Admix}-TI-DIM & ~~99.8*       & ~~99.8*       & ~~99.8*       & ~~99.9*       & 89.8          & 78.6          & 90.3          & 90.3          \\
                             & MIST-SI-TI-DIM        & ~~99.7*       & ~~99.4*       & ~~99.7*       & ~~99.9*       & \textbf{97.1} & \textbf{87.5} & \textbf{95.6} & \textbf{94.4} \\ \midrule
\multirow{4}{*}{Transformers} & \textit{Admix}        & 88.7          & 86.4          & 89.3          & 90.3          & ~~95.4*       & ~~95.5*       & ~~96.4*       & ~~98.9*       \\
                             & MIST                  & \textbf{98.4} & \textbf{97.2} & \textbf{97.4} & \textbf{98.6} & ~~98.3*       & ~~98.4*       & ~~99.1*       & ~~99.0*       \\ \cdashline{2-10}\noalign{\vskip 0.5ex}
                             & \textit{Admix}-TI-DIM & 91.7          & 88.8          & 91.1          & 92.2          & ~~97.3*       & ~~95.9*       & ~~96.0*       & ~~98.5*       \\
                             & MIST-SI-TI-DIM        & \textbf{98.4} & \textbf{97.0} & \textbf{98.0} & \textbf{98.5} & ~~98.5*       & ~~97.9*       & ~~98.9*       & ~~98.8*       \\ \bottomrule
\end{tabular}
} 
\caption{Attack success rates (\%) on eight standardly trained models by \textit{Admix} and \name, respectively.
The adversarial examples are crafted on CNN-based and Transformer-based models. * indicates white-box attacks.}
\label{tab:ensemble}
\end{table*}



\subsubsection{Evaluation on Combined Input Transformation}
SIM~\cite{lin2020nesterov}, \textit{Admix}~\cite{wang2021admix} and SSA~\cite{long2022frequency} can be combined with other input transformation-based attacks for better transferability. To evaluate \name's compatibility with these attacks, we combine \name and \name-SIM (combining \name with SIM) with TIM, DIM and TI-DIM to construct composite attacks, denoted as \name(-SI)-TIM, \name(-SI)-DIM, \name(-SI)-TI-DIM. We report the results on ResNet-18 in Fig.~\ref{fig:combined_input} and the other seven models in Appendix~\ref{appx:combined}.

We can observe that all the attacks reach nearly $100\%$ attack success rates on the ResNet-18, showing their high white-box attack effectiveness. Under the black-box setting, the adversarial transferability can be significantly enhanced when combined with various input transformation-based attacks, namely SIM, \textit{Admix}, SSA, \name and \name-SIM. In particular, \name outperforms the SOTA baseline \textit{Admix} with a clear margin of $13.33\%$, $9.22\%$ and $8.02\%$ averagely, when combined with TIM, DIM and TI-DIM, respectively. More importantly, \name-SIM, \name combined with SIM (a special case of \textit{Admix}), can further boost the transferability when integrated into the baselines, showing \name's outstanding effectiveness in crafting more transferable adversarial examples and compatibility with existing input transformation-based attacks. The results of adversarial examples crafted on the other seven models also exhibit a consistent trend, which further validates its superiority.

\subsection{Evaluation on Ensemble Model}
\label{sec:exp:ensemble}
In practice, attacking multiple models simultaneously~\cite{liu2017delving}, \aka ensemble attack, can significantly improve adversarial transferability. Existing input transformation-based attacks are often compatible with ensemble attack. To further validate \name's effectiveness, we compare \name with the best baseline \textit{Admix} and further combine them with SI-TI-DIM/TI-DIM under the ensemble setting. As shown in Sec.~\ref{sec:single_model}, the adversarial examples 
exhibit transferability across different architectures, \ie CNNs and transformers. Hence, we craft the adversarial examples on all CNNs or transformers and test them on the remaining models.

As shown in Tab.~\ref{tab:ensemble}, \name and \name-SI-TI-DIM exhibit better white-box attack success rates than \textit{Admix} and \textit{Admix}-TI-DIM, especially on transformers. This also validates the rationality of \name to separate the gradient among the mixed images. Under the black-box setting, \name achieves much better transferability than \textit{Admix} on various models, even better than \textit{Admix}-TI-DIM in most cases (except for ViT). This indicates \name's excellent compatibility with ensemble attack and further supports the superior effectiveness of \name. Notably, \name-SI-TI-DIM achieves the average attack success rate of $93.7\%$ and $98.0\%$ when taking CNNs and Transformers as the black-box surrogate models, respectively. Such outstanding performance underscores the vulnerability of a wide range of deep models to \name, which affirms the outstanding generality of \name for various input transformation-based and ensemble attacks.


\subsection{Evaluation on Defense Method}

\begin{table}[t]
\centering
\resizebox{\linewidth}{!}{
\begin{tabular}{@{}cccccccc@{}}
\toprule
Attack                & HGD & R\&P & NIPS-r3 & JPEG & FD & NRP & RS \\ \midrule
\textit{Admix}-TI-DIM & 97.3 & 95.1  &  96.1  & 98.2 & 97.7  & 89.0  & 64.4 \\
MIST-SI-TI-DIM        & \textbf{98.7} & \textbf{97.6}  & \textbf{98.0}   & \textbf{99.1} & \textbf{98.9}  & \textbf{93.3}  & \textbf{70.5} \\ \bottomrule
\end{tabular}
}
\caption{Attack success rates (\%) on seven advanced defenses by \textit{Admix}-TI-DIM and \name-SI-TI-DIM, respectively.
The adversarial examples are crafted on eight standardly trained models.}
\label{tab:defense_ensemble}
\end{table}

\begin{table}[t]
\centering
\resizebox{\linewidth}{!}{
\begin{tabular}{@{}ccccccc@{}}
\toprule
\multirow{2}{*}{Target model} & \multicolumn{6}{c}{The bound $\lambda_{min}$ for mix strength}                    \\ \cmidrule(l){2-7} 
                              & 0     & 0.1   & 0.2   & 0.3   & 0.4   & 0.5   \\ \midrule
CNN                           & 93.17 & 94.80 & 95.40 & 94.93 & 93.37 & 91.77 \\
Transformer                           & 49.50 & 53.00 & 54.15 & 54.33 & 51.20 & 47.20 \\ \bottomrule
\end{tabular}
}
\caption{Average attack success rates (\%) on three CNNs and four Transformers with various mix strengths. The adversarial examples are crafted on ResNet-18 and tested on the remaining models.}
\label{tab:ablation_lambda}
\end{table}

Until this point, the evaluations have been conducted on the standardly trained models. To mitigate the threat of adversarial examples, several adversarial defenses have been proposed. To comprehensively assess the effectiveness of \name, we consider seven advanced defense methods, namely HGD, R\&P, NIPS-r3, JPEG, FD, NRP and RS. As shown in Sec.~\ref{sec:exp:ensemble}, \textit{Admix}-TI-DIM under the ensemble setting achieves the best attack performance among the baselines. Accordingly, we adopt \textit{Admix}-TI-DIM and our \name-SI-TI-DIM to generate the adversarial examples on the eight standardly trained models and test them against these defenses.


The results are summarized in Tab.~\ref{tab:defense_ensemble}. Though \textit{Admix}-TI-DIM has exhibited superior attack performance on these defenses, our proposed \name-SI-TI-DIM consistently performs better than the baseline on all the seven adopted defenses, which achieves the attack success rate of at least $70.5\%$. On average, \name-SI-TI-DIM outperforms \textit{Admix}-TI-DIM with a clear margin of $2.6\%$. Notably, \name-SI-TI-DIM achieves an attack success rate of $93.3\%$ and $70.5\%$ on the powerful denoiser NRP or certified defense RS, respectively, exceeding those of \textit{Admix}-SI-TI-DIM by a clear margin of $4.3\%$ and $6.1\%$. These results underscore the exceptional effectiveness of our proposed \name. More critically, the superior attack performance of \name-SI-TI-DIM highlights the current limitations of existing defense mechanisms. It underscores an urgent need for the development of more robust defense strategies to ensure secure and reliable applications in practical scenarios.

\subsection{Ablation Studies}
To gain further insights into \name, we conduct parameter and ablation studies about the mix strength, number of sampled images, and the effect of mixup and random shift.

\textbf{The impact on the mix strength.}
Eq.~\ref{eq:mist} shows that the direction of \name's gradient depends on the gradients of two mixed images and mix strength $\lambda \in  \left[ \lambda_\mathrm{min}, 1-\lambda_\mathrm{min} \right]$. To ascertain an optimal interval for $\lambda$, we conduct experiments by varying $\lambda_\mathrm{min} \in [0, 0.5]$. To avoid the uncertainty resulting from the random shift, we discard the random shift operation in this ablation experiment.
As shown in Tab.~\ref{tab:ablation_lambda}, when $\lambda_\mathrm{min}$ is incrementally elevated, thereby constricting the sampling interval, the average attack success rates initially rise and subsequently decline. It suggests that the values of $\lambda$ approaching 0 or 1 yield negligible mixing potency, whereas an overly restricted sampling range curtails the diversity of the mixing process. Thus, $\lambda$ sampled near the extreme ends of the $\left[0, 1\right]$ interval and excessively narrow value intervals both constrain the improvement of attack success rates. To balance mix strength and sampling flexibility, we adopt $\lambda_\mathrm{min}=0.2$  in our experiments.


\begin{figure}[tb]
    \centering
    \includegraphics[width=0.85\linewidth]{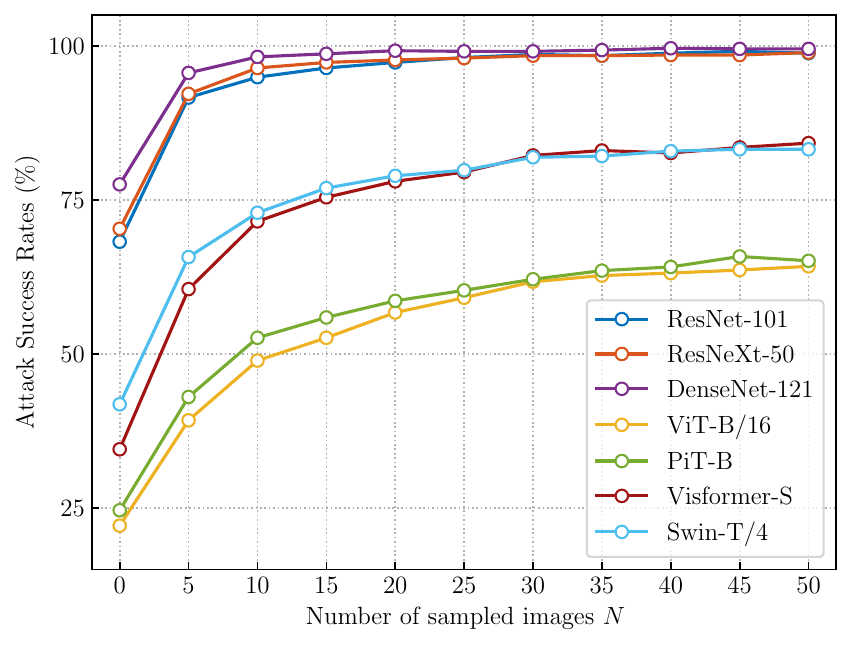}
    \caption{Attack success rates (\%) on the other seven models with the adversarial examples generated by \name on ResNet-18 using various number of sampled images $N$.}
    \label{fig:ablation_N}
\end{figure} 
\textbf{The impact on the number of sampled images.}
\name mixes several images to construct the mixed images for gradient calculation. To ascertain the optimal number of images to mix, we conduct parameter studies for $N=0\to50$ with an interval of $5$. As shown in Fig.~\ref{fig:ablation_N}, \name degenerates to MI-FGSM and is unable to exert any distinct effect when $N=0$, which achieves the poorest attack performance. When we increase the value of $N$, \name starts to be operative and the attack performance on all models stably increases, which achieves the peak around $N=30$. Further increments in $N$ lead to marginal enhancements in attack performance. Note that each mixed image needs one forward and two backward propagations, bringing extra computational costs. Thus, we set $N=30$ to balance the computational cost and attack performance in our experiments.

\begin{table*}[t]
\centering
\begin{tabular}{@{}ccccccccccc@{}}
\toprule
\multicolumn{3}{c}{Ablation} & \multicolumn{8}{c}{Target model}                                                  \\ \cmidrule(lr){1-3} \cmidrule(lr){4-11}
\textit{Admix}$^\dag$ & mixup & random shift  & ResNet-101 & ResNeXt-50 & DenseNet-121 & ViT & PiT & Visformer & Swin \\ \midrule
\ding{55} & \ding{55} & \ding{55}      & 68.2       & 70.3       & 77.5         & 22.1 & 24.6 & 34.5      & 41.8  \\
 \ding{51}  &       &      & 76.7 & 81.6 & 87.5 & 27.5 & 29.5 & 42.8 & 48.7     \\
      & \ding{51}  &                 & 93.5       & 94.9       & 97.8         & 42.1 & 44.2 & 63.8      & 66.5   \\
  & \ding{51}  &   \ding{51}   & 98.6 & 98.4 & 99.1 & 61.7 & 62.1 & 82.2 & 81.9    \\ \bottomrule
\end{tabular}

\caption{Attack success rates (\%) under different attack settings w/wo admix, mixup and random shift, respectively. The adversarial examples are crafted on ResNet-18 and tested on the other seven target models. $^\dag$ indicates \textit{Admix} without scale. }
\label{tab:ablation_shift}
\end{table*}

\textbf{The effect of mixup and random shift.}
\name randomly shifts the sampled images and mixes them with the original input image but separates the gradient in a standard mixup manner to boost the adversarial transferability. To validate whether these two components result in better transferability, we remove these components and also adopt admix without scale to generate adversarial examples on ResNet-18, then test them on the other models. The results are summarized in Tab.~\ref{tab:ablation_shift}. 
Under the black-box setting, without mixup and random shift, \name degenerates to MI-FGSM and achieves the poorest attack performance. Conversely, when we mix the image but separate the gradient, it can achieve much better transferability than MI-FGSM as well as \textit{Admix} without scale, which also mixes a small portion of images. This substantiates our hypothesis that the inherent limitations in \textit{Admix} impede optimal transferability. When we further adopt random shift, \name achieves the best attack performance, supporting the necessity of random shift and rational design of our \name.

\subsection{Further Discussion}
\begin{figure}[t]
    \centering
    \includegraphics[width=\linewidth]{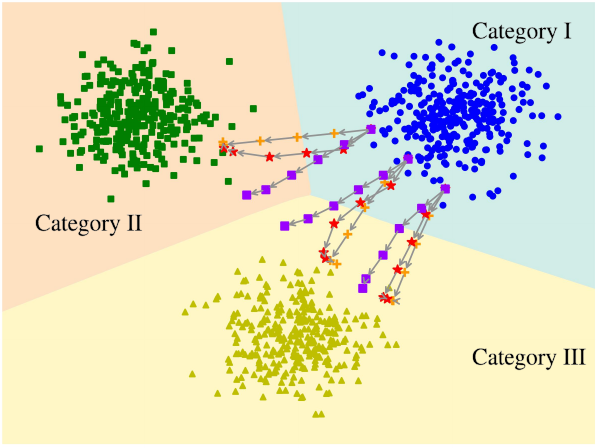}
    \caption{Visualization of optimization paths. The orange, red and purple trajectories are the optimization paths of MI-FGSM, \textit{Admix}, and \name, respectively.}
    \label{fig:vis-exp}
\end{figure}


To further elucidate the mechanisms contributing to the exceptional efficacy of \name, we visualize the optimization paths of MI-FGSM, \textit{Admix}, and \name using the toy example as in Sec.~\ref{sec:approach:rethinking}. Specifically, we randomly sample several data points in Category \RomanNum{1} and record the intermediate data points along each attack's optimization path during the generation of adversarial examples. As shown in Fig.~\ref{fig:vis-exp}, compared with MI-FGSM, \textit{Admix} generates the adversarial examples of Category \RomanNum{1} closer to the decision boundary between Category \RomanNum{2} and \RomanNum{3}. In comparison, \name further enhances this trajectory, propelling the adversarial examples closer to this critical region, resulting in better adversarial transferability. This empirical evidence validates our assumption that the adversarial examples situated at the intersection of decision boundaries for other categories are more transferable. Moreover, it provides a clear illustration of how \name successfully manipulates adversarial examples to achieve superior transferability.

\section{Conclusion}

In this work, we posit that adversarial examples located at the intersection of decision boundaries for other categories are more transferable. We identify that \textit{Admix} directs adversarial examples towards such intersections, thereby accounting for its notable efficacy. However, the constraint imposed on the incorporated image in \textit{Admix} prevents inaccurate gradient estimation but concurrently impairs its capacity to guide adversarial examples, consequently diminishing transferability. Based on this finding, we propose a novel input transformation-based attack called \textbf{M}ixing the \textbf{I}mage but \textbf{S}eparating the gradien\textbf{T} (\name). Specifically, \name combines the input image with randomly shifted images sampled from other categories without differentiating which is primary. Subsequently, it calculates the gradients \wrt these two images to update their corresponding adversarial perturbation. Extensive experiments on ImageNet dataset validate its outstanding effectiveness in boosting the transferability and remarkable compatibility with other attacks under various settings. Once again, \name certifies the superiority of adopting the information from other categories to craft more transferable adversarial examples. We hope \name could inspire more works to utilize such information for more powerful adversarial attacks and defenses.

{
    \small
    \bibliographystyle{ieeenat_fullname}
    \bibliography{main}

\begin{thebibliography}{57}
\providecommand{\natexlab}[1]{#1}
\providecommand{\url}[1]{\texttt{#1}}
\expandafter\ifx\csname urlstyle\endcsname\relax
  \providecommand{\doi}[1]{doi: #1}\else
  \providecommand{\doi}{doi: \begingroup \urlstyle{rm}\Url}\fi

\bibitem[Brendel et~al.(2018)Brendel, Rauber, and Bethge]{brendel2018decision}
Wieland Brendel, Jonas Rauber, and Matthias Bethge.
\newblock {Decision-Based Adversarial Attacks: Reliable Attacks Against
  Black-Box Machine Learning Models}.
\newblock In \emph{{Proceedings of the International Conference on Learning
  Representations}}, 2018.

\bibitem[Chen et~al.(2015)Chen, Seff, Kornhauser, and
  Xiao]{chen2015deepdriving}
Chenyi Chen, Ari Seff, Alain~L. Kornhauser, and Jianxiong Xiao.
\newblock {DeepDriving: Learning Affordance for Direct Perception in Autonomous
  Driving}.
\newblock In \emph{{Proceedings of the IEEE/CVF International Conference on
  Computer Vision}}, pages 2722--2730, 2015.

\bibitem[Chen et~al.(2021)Chen, Xie, Niu, Liu, Wei, and
  Tian]{chen2021visformer}
Zhengsu Chen, Lingxi Xie, Jianwei Niu, Xuefeng Liu, Longhui Wei, and Qi Tian.
\newblock {Visformer: The Vision-friendly Transformer}.
\newblock In \emph{{Proceedings of the IEEE/CVF International Conference on
  Computer Vision}}, pages 569--578, 2021.

\bibitem[Cohen et~al.(2019)Cohen, Rosenfeld, and Kolter]{cohen2019certified}
Jeremy Cohen, Elan Rosenfeld, and J.~Zico Kolter.
\newblock {Certified Adversarial Robustness via Randomized Smoothing}.
\newblock In \emph{{Proceedings of the International Conference on Machine
  Learning}}, pages 1310--1320, 2019.

\bibitem[Dong et~al.(2018)Dong, Liao, Pang, Su, Zhu, Hu, and
  Li]{dong2018boosting}
Yinpeng Dong, Fangzhou Liao, Tianyu Pang, Hang Su, Jun Zhu, Xiaolin Hu, and
  Jianguo Li.
\newblock {Boosting Adversarial Attacks With Momentum}.
\newblock In \emph{{Proceedings of the IEEE/CVF Conference on Computer Vision
  and Pattern Recognition}}, pages 9185--9193, 2018.

\bibitem[Dong et~al.(2019)Dong, Pang, Su, and Zhu]{dong2019evading}
Yinpeng Dong, Tianyu Pang, Hang Su, and Jun Zhu.
\newblock {Evading Defenses to Transferable Adversarial Examples by
  Translation-Invariant Attacks}.
\newblock In \emph{{Proceedings of the IEEE/CVF Conference on Computer Vision
  and Pattern Recognition}}, pages 4312--4321, 2019.

\bibitem[Dosovitskiy et~al.(2021)Dosovitskiy, Beyer, Kolesnikov, Weissenborn,
  Zhai, Unterthiner, Dehghani, Minderer, Heigold, Gelly, Uszkoreit, and
  Houlsby]{dosovitskiy2020vit}
Alexey Dosovitskiy, Lucas Beyer, Alexander Kolesnikov, Dirk Weissenborn,
  Xiaohua Zhai, Thomas Unterthiner, Mostafa Dehghani, Matthias Minderer, Georg
  Heigold, Sylvain Gelly, Jakob Uszkoreit, and Neil Houlsby.
\newblock {An Image is Worth 16x16 Words: Transformers for Image Recognition at
  Scale}.
\newblock In \emph{{Proceedings of the International Conference on Learning
  Representations}}, 2021.

\bibitem[Eykholt et~al.(2018)Eykholt, Evtimov, Fernandes, Li, Rahmati, Xiao,
  Prakash, Kohno, and Song]{eykholt2018robust}
Kevin Eykholt, Ivan Evtimov, Earlence Fernandes, Bo Li, Amir Rahmati, Chaowei
  Xiao, Atul Prakash, Tadayoshi Kohno, and Dawn Song.
\newblock {Robust Physical-World Attacks on Deep Learning Visual
  Classification}.
\newblock In \emph{{Proceedings of the IEEE/CVF Conference on Computer Vision
  and Pattern Recognition}}, pages 1625--1634, 2018.

\bibitem[Gao et~al.(2020)Gao, Zhang, Song, Liu, and Shen]{gao2020patch}
Lianli Gao, Qilong Zhang, Jingkuan Song, Xianglong Liu, and Heng~Tao Shen.
\newblock {Patch-Wise Attack for Fooling Deep Neural Network}.
\newblock In \emph{{Proceedings of the European Conference on Computer
  Vision}}, pages 307--322, 2020.

\bibitem[Ge et~al.(2023)Ge, Wang, Shang, Liu, and Liu]{ge2023boosting}
Zhijin Ge, Xiaosen Wang, Fanhua Shang, Hongying Liu, and Yuanyuan Liu.
\newblock {Boosting Adversarial Transferability by Achieving Flat Local
  Maxima}.
\newblock In \emph{Proceedings of the Advances in Neural Information Processing
  Systems}, 2023.

\bibitem[Goodfellow et~al.(2015)Goodfellow, Shlens, and
  Szegedy]{goodfellow2015explaining}
Ian~J. Goodfellow, Jonathon Shlens, and Christian Szegedy.
\newblock {Explaining and Harnessing Adversarial Examples}.
\newblock In \emph{{Proceedings of the International Conference on Learning
  Representations}}, 2015.

\bibitem[Gowal et~al.(2019)Gowal, Dvijotham, Stanforth, Bunel, Qin, Uesato,
  Arandjelovic, Mann, and Kohli]{gowal2019scalable}
Sven Gowal, Krishnamurthy Dvijotham, Robert Stanforth, Rudy Bunel, Chongli Qin,
  Jonathan Uesato, Relja Arandjelovic, Timothy~Arthur Mann, and Pushmeet Kohli.
\newblock {Scalable Verified Training for Provably Robust Image
  Classification}.
\newblock In \emph{{Proceedings of the IEEE/CVF International Conference on
  Computer Vision}}, pages 4841--4850, 2019.

\bibitem[Guo et~al.(2018)Guo, Rana, Ciss{\'{e}}, and van~der
  Maaten]{guo2018countering}
Chuan Guo, Mayank Rana, Moustapha Ciss{\'{e}}, and Laurens van~der Maaten.
\newblock {Countering Adversarial Images Using Input Transformations}.
\newblock In \emph{{Proceedings of the International Conference on Learning
  Representations}}, 2018.

\bibitem[Guo et~al.(2019)Guo, Gardner, You, Wilson, and
  Weinberger]{guo2019simple}
Chuan Guo, Jacob~R. Gardner, Yurong You, Andrew~Gordon Wilson, and Kilian~Q.
  Weinberger.
\newblock {Simple Black-box Adversarial Attacks}.
\newblock In \emph{{Proceedings of the International Conference on Machine
  Learning}}, pages 2484--2493, 2019.

\bibitem[He et~al.(2016)He, Zhang, Ren, and Sun]{he2016deep}
Kaiming He, Xiangyu Zhang, Shaoqing Ren, and Jian Sun.
\newblock {Deep Residual Learning for Image Recognition}.
\newblock In \emph{{Proceedings of the IEEE/CVF Conference on Computer Vision
  and Pattern Recognition}}, pages 770--778, 2016.

\bibitem[Heo et~al.(2021)Heo, Yun, Han, Chun, Choe, and Oh]{heo2021rethinking}
Byeongho Heo, Sangdoo Yun, Dongyoon Han, Sanghyuk Chun, Junsuk Choe, and
  Seong~Joon Oh.
\newblock {Rethinking Spatial Dimensions of Vision Transformers}.
\newblock In \emph{{Proceedings of the IEEE/CVF International Conference on
  Computer Vision}}, pages 11916--11925, 2021.

\bibitem[Huang et~al.(2017)Huang, Liu, van~der Maaten, and
  Weinberger]{huang2017densely}
Gao Huang, Zhuang Liu, Laurens van~der Maaten, and Kilian~Q. Weinberger.
\newblock {Densely Connected Convolutional Networks}.
\newblock In \emph{{Proceedings of the IEEE/CVF Conference on Computer Vision
  and Pattern Recognition}}, pages 2261--2269, 2017.

\bibitem[Ilyas et~al.(2018)Ilyas, Engstrom, Athalye, and Lin]{ilyas2018black}
Andrew Ilyas, Logan Engstrom, Anish Athalye, and Jessy Lin.
\newblock {Black-box Adversarial Attacks with Limited Queries and Information}.
\newblock In \emph{{Proceedings of the International Conference on Machine
  Learning}}, pages 2142--2151, 2018.

\bibitem[Kurakin et~al.(2017)Kurakin, Goodfellow, and
  Bengio]{kurakin2017adversarial}
Alexey Kurakin, Ian~J. Goodfellow, and Samy Bengio.
\newblock {Adversarial Examples in the Physical World}.
\newblock In \emph{{Proceedings of the International Conference on Learning
  Representations (Workshops)}}, 2017.

\bibitem[Li et~al.(2020)Li, Bai, Zhou, Xie, Zhang, and Yuille]{li2020learning}
Yingwei Li, Song Bai, Yuyin Zhou, Cihang Xie, Zhishuai Zhang, and Alan~L.
  Yuille.
\newblock {Learning Transferable Adversarial Examples via Ghost Networks}.
\newblock In \emph{{Proceedings of the AAAI Conference on Artificial
  Intelligence}}, pages 11458--11465, 2020.

\bibitem[Liao et~al.(2018)Liao, Liang, Dong, Pang, Hu, and
  Zhu]{liao2018defense}
Fangzhou Liao, Ming Liang, Yinpeng Dong, Tianyu Pang, Xiaolin Hu, and Jun Zhu.
\newblock {Defense Against Adversarial Attacks Using High-Level Representation
  Guided Denoiser}.
\newblock In \emph{{Proceedings of the IEEE/CVF Conference on Computer Vision
  and Pattern Recognition}}, pages 1778--1787, 2018.

\bibitem[Lin et~al.(2020)Lin, Song, He, Wang, and Hopcroft]{lin2020nesterov}
Jiadong Lin, Chuanbiao Song, Kun He, Liwei Wang, and John~E. Hopcroft.
\newblock {Nesterov Accelerated Gradient and Scale Invariance for Adversarial
  Attacks}.
\newblock In \emph{{Proceedings of the International Conference on Learning
  Representations}}, 2020.

\bibitem[Liu et~al.(2017)Liu, Chen, Liu, and Song]{liu2017delving}
Yanpei Liu, Xinyun Chen, Chang Liu, and Dawn Song.
\newblock {Delving into Transferable Adversarial Examples and Black-box
  Attacks}.
\newblock In \emph{{Proceedings of the International Conference on Learning
  Representations}}, 2017.

\bibitem[Liu et~al.(2019)Liu, Liu, Liu, Xu, Lin, Wang, and Wen]{liu2019feature}
Zihao Liu, Qi Liu, Tao Liu, Nuo Xu, Xue Lin, Yanzhi Wang, and Wujie Wen.
\newblock {Feature Distillation: DNN-Oriented{ JPEG} Compression Against
  Adversarial Examples}.
\newblock In \emph{{Proceedings of the IEEE/CVF Conference on Computer Vision
  and Pattern Recognition}}, pages 860--868, 2019.

\bibitem[Liu et~al.(2021)Liu, Lin, Cao, Hu, Wei, Zhang, Lin, and
  Guo]{liu2021swin}
Ze Liu, Yutong Lin, Yue Cao, Han Hu, Yixuan Wei, Zheng Zhang, Stephen Lin, and
  Baining Guo.
\newblock {Swin Transformer: Hierarchical Vision Transformer Using Shifted
  Windows}.
\newblock In \emph{{Proceedings of the IEEE/CVF International Conference on
  Computer Vision}}, pages 9992--10002, 2021.

\bibitem[Long et~al.(2022)Long, Zhang, Zeng, Gao, Liu, Zhang, and
  Song]{long2022frequency}
Yuyang Long, Qilong Zhang, Boheng Zeng, Lianli Gao, Xianglong Liu, Jian Zhang,
  and Jingkuan Song.
\newblock Frequency domain model augmentation for adversarial attack.
\newblock In \emph{Proceedings of the European Conference on Computer Vision},
  pages 549--566, 2022.

\bibitem[Madry et~al.(2018)Madry, Makelov, Schmidt, Tsipras, and
  Vladu]{madry2018towards}
Aleksander Madry, Aleksandar Makelov, Ludwig Schmidt, Dimitris Tsipras, and
  Adrian Vladu.
\newblock {Towards Deep Learning Models Resistant to Adversarial Attacks}.
\newblock In \emph{{Proceedings of the International Conference on Learning
  Representations}}, 2018.

\bibitem[Naseer et~al.(2020)Naseer, Khan, Hayat, Khan, and
  Porikli]{naseer2020a}
Muzammal Naseer, Salman~H. Khan, Munawar Hayat, Fahad~Shahbaz Khan, and Fatih
  Porikli.
\newblock {A Self-supervised Approach for Adversarial Robustness}.
\newblock In \emph{{Proceedings of the IEEE/CVF Conference on Computer Vision
  and Pattern Recognition}}, pages 259--268, 2020.

\bibitem[Naseer et~al.(2021)Naseer, Khan, Hayat, Khan, and
  Porikli]{naseer2021generating}
Muzammal Naseer, Salman~H. Khan, Munawar Hayat, Fahad~Shahbaz Khan, and Fatih
  Porikli.
\newblock {On Generating Transferable Targeted Perturbations}.
\newblock In \emph{{Proceedings of the IEEE/CVF International Conference on
  Computer Vision}}, pages 7688--7697, 2021.

\bibitem[Poursaeed et~al.(2018)Poursaeed, Katsman, Gao, and
  Belongie]{poursaeed2018generative}
Omid Poursaeed, Isay Katsman, Bicheng Gao, and Serge~J. Belongie.
\newblock {Generative Adversarial Perturbations}.
\newblock In \emph{{Proceedings of the IEEE/CVF Conference on Computer Vision
  and Pattern Recognition}}, pages 4422--4431, 2018.

\bibitem[Russakovsky et~al.(2015)Russakovsky, Deng, Su, Krause, Satheesh, Ma,
  Huang, Karpathy, Khosla, Bernstein, Berg, and Fei{-}Fei]{Russa2015imagenet}
Olga Russakovsky, Jia Deng, Hao Su, Jonathan Krause, Sanjeev Satheesh, Sean Ma,
  Zhiheng Huang, Andrej Karpathy, Aditya Khosla, Michael~S. Bernstein,
  Alexander~C. Berg, and Li Fei{-}Fei.
\newblock {ImageNet Large Scale Visual Recognition Challenge}.
\newblock \emph{{International journal of computer vision}}, 115\penalty0
  (3):\penalty0 211--252, 2015.

\bibitem[Simonyan and Zisserman(2015)]{simonyan2015very}
Karen Simonyan and Andrew Zisserman.
\newblock {Very Deep Convolutional Networks for Large-Scale Image Recognition}.
\newblock In \emph{{Proceedings of the International Conference on Learning
  Representations}}, 2015.

\bibitem[Szegedy et~al.(2014)Szegedy, Zaremba, Sutskever, Bruna, Erhan,
  Goodfellow, and Fergus]{szegedy2014intriguing}
Christian Szegedy, Wojciech Zaremba, Ilya Sutskever, Joan Bruna, Dumitru Erhan,
  Ian~J. Goodfellow, and Rob Fergus.
\newblock {Intriguing Properties of Neural Networks}.
\newblock In \emph{{Proceedings of the International Conference on Learning
  Representations}}, 2014.

\bibitem[Tram{\`{e}}r et~al.(2018)Tram{\`{e}}r, Kurakin, Papernot, Goodfellow,
  Boneh, and McDaniel]{tramer2018ensemble}
Florian Tram{\`{e}}r, Alexey Kurakin, Nicolas Papernot, Ian~J. Goodfellow, Dan
  Boneh, and Patrick~D. McDaniel.
\newblock {Ensemble Adversarial Training: Attacks and Defenses}.
\newblock In \emph{{Proceedings of the International Conference on Learning
  Representations}}, 2018.

\bibitem[Wang et~al.(2018)Wang, Wang, Zhou, Ji, Gong, Zhou, Li, and
  Liu]{wang2018cosface}
Hao Wang, Yitong Wang, Zheng Zhou, Xing Ji, Dihong Gong, Jingchao Zhou, Zhifeng
  Li, and Wei Liu.
\newblock {CosFace: Large Margin Cosine Loss for Deep Face Recognition}.
\newblock In \emph{{Proceedings of the IEEE/CVF Conference on Computer Vision
  and Pattern Recognition}}, pages 5265--5274, 2018.

\bibitem[Wang et~al.(2023{\natexlab{a}})Wang, He, Wang, and
  Wang]{wang2023boosting}
Kunyu Wang, Xuanran He, Wenxuan Wang, and Xiaosen Wang.
\newblock {Boosting Adversarial Transferability by Block Shuffle and Rotation}.
\newblock \emph{arXiv preprint arXiv:2308.10299}, 2023{\natexlab{a}}.

\bibitem[Wang and He(2021)]{wang2021enhancing}
Xiaosen Wang and Kun He.
\newblock {Enhancing the Transferability of Adversarial Attacks Through
  Variance Tuning}.
\newblock In \emph{{Proceedings of the IEEE/CVF Conference on Computer Vision
  and Pattern Recognition}}, pages 1924--1933, 2021.

\bibitem[Wang et~al.(2021{\natexlab{a}})Wang, He, Wang, and He]{wang2021admix}
Xiaosen Wang, Xuanran He, Jingdong Wang, and Kun He.
\newblock {Admix: Enhancing the Transferability of Adversarial Attacks}.
\newblock In \emph{{Proceedings of the IEEE/CVF International Conference on
  Computer Vision}}, pages 16138--16147, 2021{\natexlab{a}}.

\bibitem[Wang et~al.(2021{\natexlab{b}})Wang, Lin, Hu, Wang, and
  He]{wang2021boosting}
Xiaosen Wang, Jiadong Lin, Han Hu, Jingdong Wang, and Kun He.
\newblock {Boosting Adversarial Transferability through Enhanced Momentum}.
\newblock In \emph{{Proceedings of the British Machine Vision Conference}},
  page 272, 2021{\natexlab{b}}.

\bibitem[Wang et~al.(2022)Wang, Zhang, Tong, Gong, He, Li, and
  Liu]{wang2022triangle}
Xiaosen Wang, Zeliang Zhang, Kangheng Tong, Dihong Gong, Kun He, Zhifeng Li,
  and Wei Liu.
\newblock {Triangle Attack:{ A} Query-Efficient Decision-Based Adversarial
  Attack}.
\newblock In \emph{{Proceedings of the European Conference on Computer
  Vision}}, pages 156--174, 2022.

\bibitem[Wang et~al.(2023{\natexlab{b}})Wang, Tong, and He]{wang2023rethinking}
Xiaosen Wang, Kangheng Tong, and Kun He.
\newblock {Rethinking the Backward Propagation for Adversarial
  Transferability}.
\newblock In \emph{Proceedings of the Advances in Neural Information Processing
  Systems}, 2023{\natexlab{b}}.

\bibitem[Wang et~al.(2023{\natexlab{c}})Wang, Zhang, and
  Zhang]{wang2023structure}
Xiaosen Wang, Zeliang Zhang, and Jianping Zhang.
\newblock {Structure Invariant Transformation for better Adversarial
  Transferability}.
\newblock In \emph{Proceedings of the IEEE/CVF International Conference on
  Computer Vision}, pages 4607--4619, 2023{\natexlab{c}}.

\bibitem[Wei et~al.(2019)Wei, Liang, Chen, and Cao]{wei2019transferable}
Xingxing Wei, Siyuan Liang, Ning Chen, and Xiaochun Cao.
\newblock {Transferable Adversarial Attacks for Image and Video Object
  Detection}.
\newblock In \emph{{Proceedings of the International Joint Conference on
  Artificial Intelligence}}, pages 954--960, 2019.

\bibitem[Wen et~al.(2016)Wen, Zhang, Li, and Qiao]{wen2016discriminative}
Yandong Wen, Kaipeng Zhang, Zhifeng Li, and Yu Qiao.
\newblock {A Discriminative Feature Learning Approach for Deep Face
  Recognition}.
\newblock In \emph{{Proceedings of the European Conference on Computer
  Vision}}, pages 499--515, 2016.

\bibitem[Wong et~al.(2020)Wong, Rice, and Kolter]{wong2020fast}
Eric Wong, Leslie Rice, and J.~Zico Kolter.
\newblock {Fast is Better than Free: Revisiting Adversarial Training}.
\newblock In \emph{{Proceedings of the International Conference on Learning
  Representations}}, 2020.

\bibitem[Wu et~al.(2020)Wu, Wang, Xia, Bailey, and Ma]{wu2020skip}
Dongxian Wu, Yisen Wang, Shu{-}Tao Xia, James Bailey, and Xingjun Ma.
\newblock {Skip Connections Matter: On the Transferability of Adversarial
  Examples Generated with ResNets}.
\newblock In \emph{{Proceedings of the International Conference on Learning
  Representations}}, 2020.

\bibitem[Wu et~al.(2021)Wu, Su, Lyu, and King]{wu2021improving}
Weibin Wu, Yuxin Su, Michael~R. Lyu, and Irwin King.
\newblock {Improving the Transferability of Adversarial Samples With
  Adversarial Transformations}.
\newblock In \emph{{Proceedings of the IEEE/CVF Conference on Computer Vision
  and Pattern Recognition}}, pages 9024--9033, 2021.

\bibitem[Xie et~al.(2018)Xie, Wang, Zhang, Ren, and Yuille]{xie2018mitigating}
Cihang Xie, Jianyu Wang, Zhishuai Zhang, Zhou Ren, and Alan~L. Yuille.
\newblock {Mitigating Adversarial Effects Through Randomization}.
\newblock In \emph{{Proceedings of the International Conference on Learning
  Representations}}, 2018.

\bibitem[Xie et~al.(2019)Xie, Zhang, Zhou, Bai, Wang, Ren, and
  Yuille]{xie2019improving}
Cihang Xie, Zhishuai Zhang, Yuyin Zhou, Song Bai, Jianyu Wang, Zhou Ren, and
  Alan~L. Yuille.
\newblock {Improving Transferability of Adversarial Examples With Input
  Diversity}.
\newblock In \emph{{Proceedings of the IEEE/CVF Conference on Computer Vision
  and Pattern Recognition}}, pages 2730--2739, 2019.

\bibitem[Xie et~al.(2017)Xie, Girshick, Doll{\'{a}}r, Tu, and
  He]{xie2017aggregated}
Saining Xie, Ross~B. Girshick, Piotr Doll{\'{a}}r, Zhuowen Tu, and Kaiming He.
\newblock {Aggregated Residual Transformations for Deep Neural Networks}.
\newblock In \emph{{Proceedings of the IEEE/CVF Conference on Computer Vision
  and Pattern Recognition}}, pages 5987--5995, 2017.

\bibitem[Xiong et~al.(2022)Xiong, Lin, Zhang, Hopcroft, and
  He]{xiong2022stochastic}
Yifeng Xiong, Jiadong Lin, Min Zhang, John~E. Hopcroft, and Kun He.
\newblock {Stochastic Variance Reduced Ensemble Adversarial Attack for Boosting
  the Adversarial Transferability}.
\newblock In \emph{{Proceedings of the IEEE/CVF Conference on Computer Vision
  and Pattern Recognition}}, pages 14963--14972, 2022.

\bibitem[Yuan et~al.(2022)Yuan, Zhang, and Shan]{yuan2022adaptive}
Zheng Yuan, Jie Zhang, and Shiguang Shan.
\newblock {Adaptive Image Transformations for Transfer-Based Adversarial
  Attack}.
\newblock In \emph{{Proceedings of the European Conference on Computer
  Vision}}, pages 1--17, 2022.

\bibitem[Zhang et~al.(2018{\natexlab{a}})Zhang, Ciss{\'{e}}, Dauphin, and
  Lopez{-}Paz]{zhang2018mixup}
Hongyi Zhang, Moustapha Ciss{\'{e}}, Yann~N. Dauphin, and David Lopez{-}Paz.
\newblock {Mixup: Beyond Empirical Risk Minimization}.
\newblock In \emph{{Proceedings of the International Conference on Learning
  Representations}}, 2018{\natexlab{a}}.

\bibitem[Zhang et~al.(2018{\natexlab{b}})Zhang, Weng, Chen, Hsieh, and
  Daniel]{zhang2018efficient}
Huan Zhang, Tsui{-}Wei Weng, Pin{-}Yu Chen, Cho{-}Jui Hsieh, and Luca Daniel.
\newblock {Efficient Neural Network Robustness Certification with General
  Activation Functions}.
\newblock In \emph{{Proceedings of the Advances in Neural Information
  Processing Systems}}, pages 4944--4953, 2018{\natexlab{b}}.

\bibitem[Zhang et~al.(2022{\natexlab{a}})Zhang, Wu, Huang, Huang, Wang, Su, and
  Lyu]{zhang2022improving}
Jianping Zhang, Weibin Wu, Jen{-}tse Huang, Yizhan Huang, Wenxuan Wang, Yuxin
  Su, and Michael~R. Lyu.
\newblock {Improving Adversarial Transferability via Neuron Attribution-based
  Attacks}.
\newblock In \emph{{Proceedings of the IEEE/CVF Conference on Computer Vision
  and Pattern Recognition}}, pages 14973--14982, 2022{\natexlab{a}}.

\bibitem[Zhang et~al.(2022{\natexlab{b}})Zhang, Tan, Chen, Liu, Zhang, and
  Li]{zhang2022enhancing}
Yaoyuan Zhang, Yu{-}an Tan, Tian Chen, Xinrui Liu, Quanxin Zhang, and Yuanzhang
  Li.
\newblock {Enhancing the Transferability of Adversarial Examples with Random
  Patch}.
\newblock In \emph{{Proceedings of the International Joint Conference on
  Artificial Intelligence}}, pages 1672--1678, 2022{\natexlab{b}}.

\bibitem[Zhou et~al.(2018)Zhou, Hou, Chen, Tang, Huang, Gan, and
  Yang]{zhou2018transferable}
Wen Zhou, Xin Hou, Yongjun Chen, Mengyun Tang, Xiangqi Huang, Xiang Gan, and
  Yong Yang.
\newblock {Transferable Adversarial Perturbations}.
\newblock In \emph{{Proceedings of the European Conference on Computer
  Vision}}, pages 471--486, 2018.

\end{thebibliography}
}

\clearpage
\appendix

\section{Additional Evaluations on combined input transformation}
\label{appx:combined}

\begin{figure*}[t]
    \centering
    \vspace{-1em}
    \includegraphics[width=\linewidth]{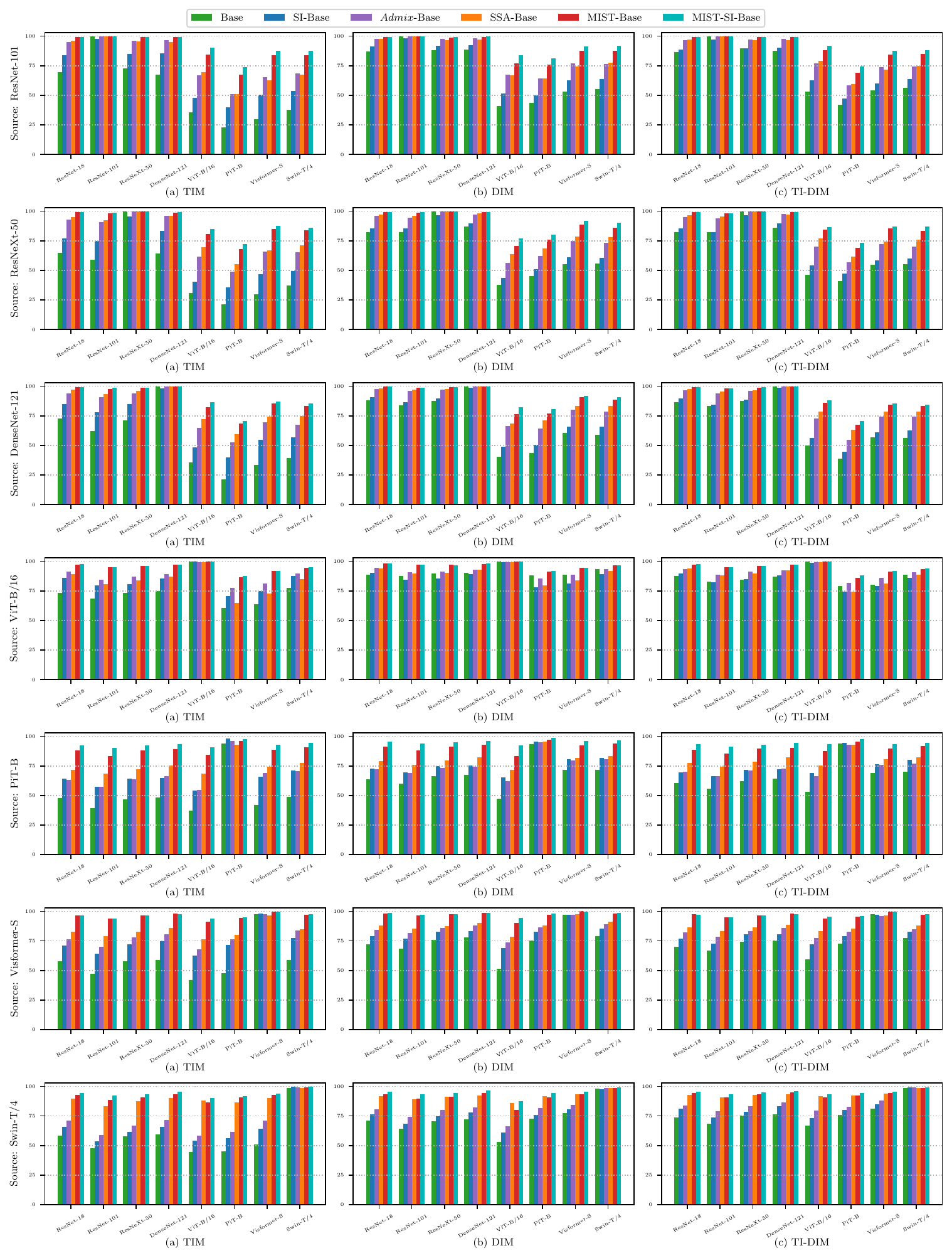}
    \vspace{-2.2em}
    \caption{Attack success rates (\%) on eight standardly trained models with the adversarial examples generated on the seven source models using different combined input transformations.}
    \label{fig:combined_input_appendix}
\end{figure*}

In Sec.~\ref{sec:single_model}, we provide the attack results of the adversarial examples generated by various combined input transformation-based attacks on ResNet-18. To further validate the compatibility of \name, we generate adversarial examples on the remaining seven source models, \ie, ResNet-101, ResNeXt-50, DenseNet-121, ViT, PiT, Visformer as well as Swin and test them on the other models. 

The results are summarized in Fig.~\ref{fig:combined_input_appendix}. Under the white-box setting, the baselines can achieve the attack success rate of nearly $100\%$ on five models (\ie, ResNet-101, ResNeXt-50, DenseNet-121, ViT, Swin) but exhibit slightly poor performance on the other two models (\ie, PiT, Visformer). On the contrary, our proposed \name consistently achieves the attack success rate of nearly $100\%$ on all the models when combined with various input transformations, showing its stability and generality for different architectures. Under the black-box setting, \name combined with various input transformations achieves significantly better attack performance than that combined with other baselines on all the models. Also, when integrating \name-SIM into these input transformation-based attacks, we could obtain the best attack performance under the same setting. These results are consistent with that of adversarial examples generated on ResNet-18 as shown in Sec.~\ref{sec:single_model}. Once again, these superior results demonstrate \name's outstanding effectiveness in boosting adversarial attacks and generality for different architectures.

\end{document}